\documentclass{article}
\usepackage[left=2.54cm, right=2.54cm, top=2.54cm, bottom = 2.54cm]{geometry}
\usepackage{url}
\usepackage{graphicx}
\usepackage{amsmath}
\usepackage[normalem]{ulem}
\useunder{\uline}{\ul}{}
\usepackage{pgfplots}
\usepackage{scrextend}
\usepackage{url}
\pgfplotsset{width=8cm,compat=1.9}

\usepackage{setspace}
\begin{document}\sloppy

\title{Exploring the Performance of Deep Residual Networks in Crazyhouse Chess}
\date{}
\author{Gordon Chi\\gsychi@stanford.edu}

\maketitle

\begin{center}
\url{https://github.com/FTdiscovery/64CrazyhouseDeepLearning}
\end{center}

\begin{addmargin}[5.5em]{5.5em}
\begin{center}
\section*{Abstract}
\end{center}
Crazyhouse is a chess variant that incorporates all of the classical chess rules, but allows users to drop pieces captured from the opponent as a normal move. Until 2018, all competitive computer engines for this board game made use of an alpha-beta pruning algorithm with a hand-crafted evaluation function for each position. Previous machine learning-based algorithms for just regular chess, such as NeuroChess \cite{thrun1995learning} and Giraffe \cite{lai2015giraffe}, took hand-crafted evaluation features as input rather than a raw board representation. More recent projects, such as AlphaZero, reached massive success but required massive computational resources in order to reach its final strength \cite{SilverHuangEtAl16nature}\cite{silver2017mastering1}. 
\\ \\
This paper describes the development of SixtyFour, an engine designed to compete in the chess variant of Crazyhouse with limited hardware. This specific variant poses a multitude of significant challenges due to its large branching factor, state-space complexity, and the multiple move types a player can make. We propose the novel creation of a neural network-based evaluation function for Crazyhouse. More importantly, we evaluate the effectiveness of an ensemble model, which allows the training time and datasets to be easily distributed on regular CPU hardware commodity. Early versions of the network have attained a playing level comparable to a strong amateur on online servers.
\end{addmargin}

\section{Introduction}
\subsection{Background and Motivation}

Alpha-beta pruning, although the predominant search algorithm for current board game AIs, suffers from a few drawbacks. The main problem is known as the horizon effect, and the effects are exacerbated when hardware is insufficient. If a game engine is unable to calculate the end outcome of the game, it will output an overly optimistic evaluation of a final position, as well as miss long term weaknesses that it assumes will not be targeted. More importantly, the performance of alpha-beta pruning engines is completely dependent on the domain of human knowledge. An evaluation function can only be crafted and refined based on our knowledge of important features; a use of a neural network evaluation function is too slow to be implemented alongside a minimax algorithm. Additionally, the performance speed of the alpha-beta pruning approach is dependent on the ordering of moves. Without much knowledge of a new game, the search algorithm will not be able to efficiently prune off 'bad' moves in any given position.

Monte Carlo Tree Search is an alternative method to prune and traverse the search tree of board positions. Each playout of the Monte Carlo Tree Search starts off by analyzing the most promising move in a given position, and playing out from the position until the game ends. The game result is then used to update the evaluation of each position. This technique, when paired with a neural network, is especially useful for games with large branching factors. The network predictions are capable of guiding the tree to explore only the most promising moves, regardless of the number of legal moves in a position. We believe this is the key to successful AIs for more complex games.

\subsection{Proposed Method}

The aforementioned weaknesses of alpha-beta pruning are evident in all Crazyhouse-playing agents. Crazyhouse is a chess variant that incorporates all of the classical chess rules, but allows users to drop pieces captured from the opponent as a normal move. Pieces may be dropped onto any empty square, with the exception of pawns, which can only be dropped between the second and seventh rank. Consequently, the number of legal moves is up to three times more than traditional chess, making it a considerably more complex game than just traditional chess. At the same time, the percentage rate of draws in Crazyhouse is near zero percent, which implies that there only exists a few correct moves in a position. For that reason, Crazyhouse engines have always been a longstanding challenge for programmers. An exhaustive search is both infeasible and strategically lacking.

A neural-network based engine has yet to be attempted for the game of Crazyhouse. However, the features of the game makes this a sensible approach. A successful engine must acquire a strong understanding of king safety, initiative, and attacking potential in any position. Both concepts are abstract in nature and are hard to craft by hand; patterns discovered by a neural network would emulate human play without any explicit understanding of features. Tactical motifs and mating patterns, on another hand, can only be verified through calculation in traditional alpha-beta engines. However, this is fundamentally different from how humans think. We desire for our final agent to intuitively move away from danger and execute knight forks simply by looking at the positional features. 

When compared to Chess, there are many more board configurations, meaning that network predictions have to make sense for an even wider range of possibilities. Predictions also have to be much more accurate, as the tactical nature of the game makes even one mistake in fifty moves unforgivable. The game is also comparable to the difficulty of Go, as there are many more piece types and move possibilities (one may place or relocate a piece). Ultimately, this requires more information to be encoded in both the input and output vector. For that reason, the training approach, the representation of the board and move actions, and the neural network architecture are all of utmost importance.

Taking on the challenge, this paper describes the process of training a supervised policy model made to 'emulate' the play of a strong amateur online. Then, using this a base checkpoint, we improve the network by reinforcing the network with self-play games generated with the help of the Monte Carlo Tree Search. We hope this approach will be enough to create a faster, more efficient Crazyhouse program. 

\section{Related Works}

Creating general algorithms to successfully outperform the elites of checkers, chess, Go, and similar board games have been longstanding goals for computer science in the last century. Mastering such a skill, as many had once believed, required an edge of human intuition and creativity that brute force, number-crunching algorithms were unable to excel in. This theory was quickly disproved by the successes in multiple man versus machine tournaments. In 1990, Chinook became the first computer program to win a world champion title against humans, and came second to Marion Tinsley in the US Nationals. Then, just six years later, Deep Blue became the first program to defeat a world champion in a game of classical chess; this was followed up by a 3.5-2.5 rematch victory over Garry Kasparov in a 6 game classical match-up in New York, 1997. Since then, researchers have focused their efforts into exploring the success of new algorithms, and into creating a superhuman agent for the more complex game of Go.

\subsection{Neural Network Engines}

It was once believed that with the advancement of hardware, alpha-beta pruning search algorithms would be sufficient to defeat professional players of the more complex game of Go. Feng Hsiung-Hsu, the architect of Deep Blue, once predicted that "...a world-champion-level Go machine can be build within 10 years, based upon the same method of intensive analysis--brute force, basically--that Deep Blue employed for chess..."\cite{feng}. Nevertheless, there have been different attempts that incorporate neural networks into working chess and go engines. 

NeuroChess \cite{thrun1995learning} uses temporal difference learning and takes in 120,000 grandmaster games as input to tune an evaluation function. It is capable of beating GNUChess 13\% of the time when both engines are tuned to the same search depth of 2, 3, or 4. Giraffe Chess by Lai \cite{lai2015giraffe} proposes the use of the TDLeaf($\lambda$) and a probability-based search framework in 2015; it attained a level of play comparable to a FIDE International Master (2400).

By far the most famous deep learning-based engine is AlphaGo, the first machine to defeat a top professional human Go player. \cite{SilverHuangEtAl16nature} Not only did it evaluate positions in a more human-like manner, but it also eliminated the need to craft a complex evaluation function for end positions. This project was modified in 2017 to learn entirely through reinforcement learning, and was capable of defeating top open-source engines for Chess and Shogi as well after tens and millions of self-play games.

\subsection{Past Crazyhouse Engines}

There have been successful attempts to create a superhuman Crazyhouse engine. Prior to the creation of multi-variant Stockfish, Sunsetter and Sjeng had attained a playing level of above 2500 ELO. Each engine utilizes an alpha-beta pruning search and unique approaches to the evaluation of king safety and piece drops. 
\\ \\
At around the same time of this project, a deep-learning based engine named CrazyAra was individually developed by Johannes Czech et al.; their work can be found \url{https://github.com/queensgambit/crazyara}. Like SixtyFour, their final algorithm obtained a strong understanding of the game and was even able to defeat Stockfish in 20 minute time controls.

\section{Supervised Learning}

We begin by developing a supervised learning model capable of imitating moves of strong amateur players based on limited knowledge. Our Crazyhouse engine is only given perfect knowledge of the game rules and, unlike previous projects such as NeuroChess \cite{thrun1995learning}and Giraffe \cite{lai2015giraffe}, is not given any access to more advanced features of the board. 

\subsection{Board Representation}

The input for the neural network is a basic raw representation of the board. In total, there are 960 (15 $\times$ 8 $\times$ 8) total inputs, first consisting of 12 8 $\times$ 8 planes (pawn, knight, bishop, rook, queen, and king placement for each player) that encode for the position of pieces on the board. Next, we add a binary 8 $\times$ 8 plane that denotes if any current non-pawn piece is a promoted piece. This is an important addition, as capturing a promoting piece results will only yield a pawn in the pocket of the capturing player (as opposed to the full promoted value).

For the next plane, the first 62 inputs encode full information about each player's pocket and the pieces they are allowed to drop. In the most extreme scenarios, a player can have up to 16 pawns, 4 knights, 4 bishops, 4 rooks, and 2 queens ready to be dropped. This is a better alternative to representing the number of captured pieces as a scaled constant, as a strategic plan may be completely different depending on the number of captive piece. The difference of one captive pawn, for example, can only be distinct if every single pawn is represented as a different input variable. The last two inputs denote the turn of the player.

For the final plane, the first four inputs encode the rights of kingside castling and queenside castling for both players. The next two inputs encode whether the position has been seen before either once or twice (any more would yield a threefold repetition, which is automatically a draw). The other 58 inputs are zeros, but are added in our input representation regardless in order to preserve the 8 $\times$ 8 board shape.

\begin{table}
\centering
\begin{tabular}{ |p{5cm}|p{4cm}|}
\hline
\multicolumn{2}{|c|}{\textbf{Board Representation }} \\
\hline
\textbf{Feature} & \textbf{Number of Inputs} \\
\hline
P1 Piece & 6 $\times$ 8 $\times$ 8 = 384 \\
P2 Piece & 6 $\times$ 8 $\times$ 8 = 384 \\
Promoted Pawn Plane & 1 $\times$ 8 $\times$ 8 = 64 \\
P1 Captive Pawns & 16\\
P2 Captive Pawns & 16 \\
P1 Captive Knights & 4\\
P2 Captive Knights & 4 \\
P1 Captive Bishops & 4\\
P2 Captive Bishops & 4 \\
P1 Captive Rooks & 4\\
P2 Captive Rooks & 4 \\
P1 Captive Queen & 2\\
P2 Captive Queen & 2 \\
P1 to Move & 1 \\
P2 to Move & 1 \\
P1 Kingside Castling Rights & 1 \\
P2 Kingside Castling Rights & 1 \\
P1 Queenside Castling Rights & 1 \\
P2 Queenside Castling Rights & 1 \\
Position Repetition Information & 2 \\
Blank & 58 \\
\hline
\textbf{Total} & \textbf{960} \\
\hline
\end{tabular}
\textbf{\caption{Input Representation of the Board}}
\end{table}

\newpage

\subsection{Output Representation}

\subsubsection{Move Representation} 

AlphaZero proposed a method of representing all chess moves in 8 $\times$ 8 planes that encode for all possible linear movements. However, such a method is unnecessary, as it encodes for many illegal moves; this is due to the fact that a piece may never move to a region out of the board. For example, a piece on the center of the board will never be able to move seven rows in any direction, whereas this is not the case for a piece on the edge of the board; similarly, a pawn may never be dropped in the first or the eight rank of the board. (This can be better visualized in the figure below.) SixtyFour improves upon this approach and utilises a scaled one-hot vector of length 2308 to encode all possible moves in the game of Crazyhouse.

\begin{figure}[!htb]
\centering
\includegraphics[width=11cm]{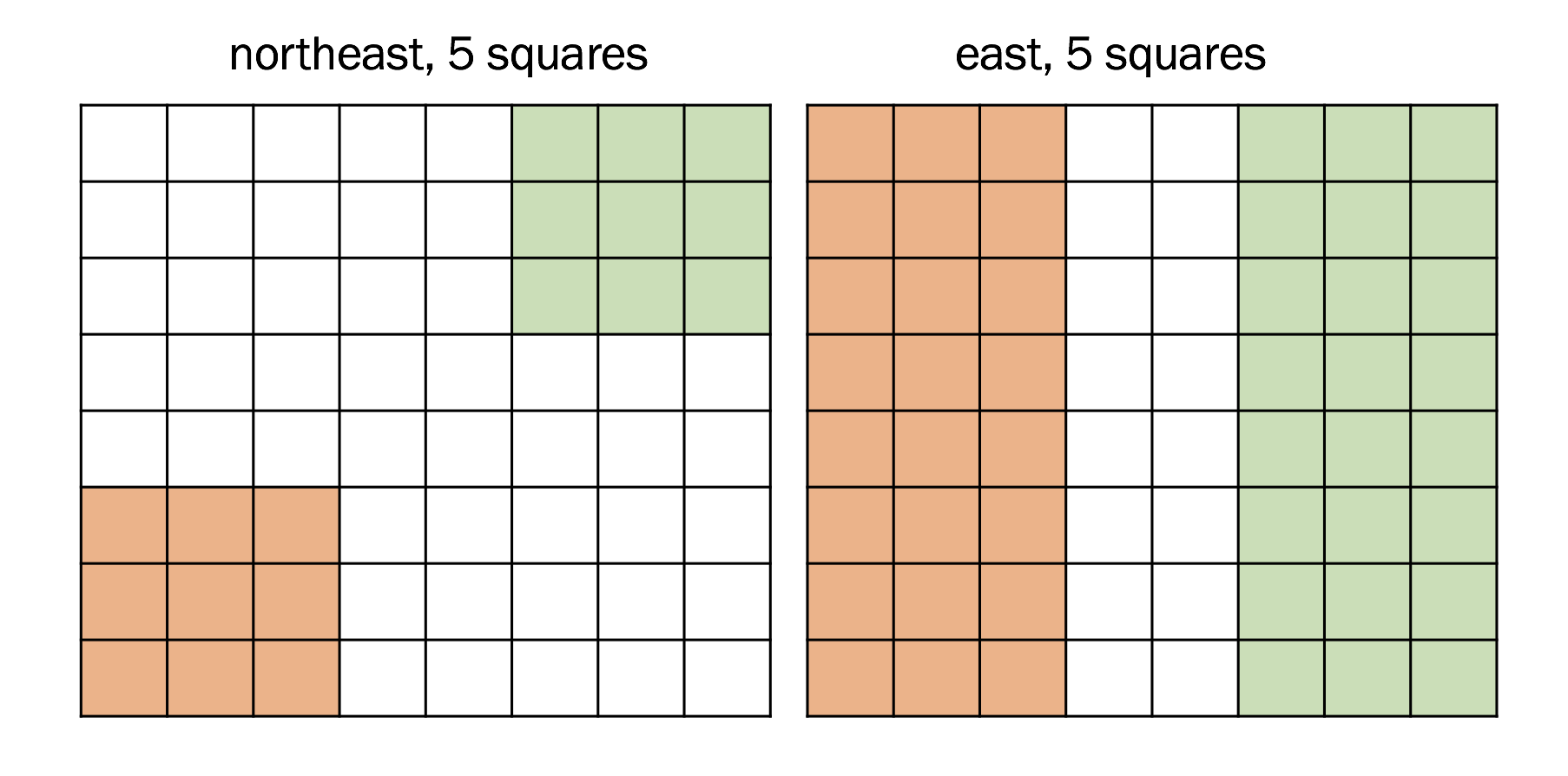}
\textbf{\caption{Consider the two above grids representing a chessboard. In the left, we consider all five-square linear moves in the northeastern (NE) direction. In the right, we consider all five-square linear moves in the eastern (E) direction. For these two types of moves, the piece must reside in the orange region and move to the green region. This means that for all five-square NE moves, only 9 out of 64 squares correspond to legal moves; for all five-square E moves, only 24 out of 64 squares correspond to legal moves. Thus, it is inefficient to represent these moves in two $8 \times 8$ planes. We need only 33 inputs to represent the moves in the figure.}}

\label{action}
\end{figure} 
\vspace{0.5cm}
\noindent In total, there are four types of moves in the game of Crazyhouse.
\\ \\
\textbf{Piece Drops:} The first 304 inputs (5 $\times$ 64 - 16) represent the possible squares in which a given piece can be placed (pawn, knight, bishop, rook, queen).
\\ \\
\textbf{Linear Moves:} To begin, we construct 56 binary planes to represent all possible linear moves applicable to a pawn, bishop, rook, queen, and king, where a value of 1 is denoted for the square that the piece will land on (and 0 for every other square). There are 8 possible directions (N, NE, E, SE, S, SW, W, NW) and, for each direction, there are 7 planes for how far a piece can move (1-7). We then remove all redundant squares as shown in Figure 1, with the exception of length one diagonal moves. This removes a total of 1172 entries from diagonally linear moves and 896 entries from horizontal/vertical moves, resulting in a total of 1516 inputs.
\\ \\
\textbf{Knight Moves:} The final 8 planes map all possible knight moves, with each plane denoting a different direction (i.e. two squares forwards and one square to the right). We then remove all unnecessarily mapped squares on the first, second, seventh and eighth ranks depending on the direction of the knight move, resulting in a total of 416 (8 $\times$ 64 - (16 $\times$ 4 + 8 $\times$ 4)) inputs.
\\ \\
\textbf{Underpromotions:} Any linear move of a pawn that reaches the final rank is automatically promoted to a queen under our current representation. However, since tempo is of the utmost importance in Crazyhouse, underpromotions to a knight/bishops/rooks can often be advantageous and must be represented in some form. 
\\
An underpromotion to each piece is represented in 3 $\times$ 8 inputs, as there are eight promotion squares and three ways to arrive at that promotion square (a left diagonal capture, a right diagonal capture, or a normal pawn move). Since there are three possible pieces to underpromote to, our underpromotions are represented in a total of 72 inputs.
\\ \\ 
Thus, our output is an array of length 304 + 1516 + 416 + 72 = 2308. Only one entry has a non-zero value.

\vspace{0.5cm}
\subsubsection{Policy Representation should depend on game result}
In 2016, DeepMind proposed that all move representations, regardless of their win rate, should remain a one-hot encoded vector so long as the supervised dataset chooses to play the move. For SixtyFour, we scale the magnitude of the one-hot encoded vectors depending on the outcome of the game. Vectors of moves played by the winning side will remain unchanged; vectors of moves played by the losing side will be multiplied by 0.1; vectors of moves played in a draw will be multiplied by 0.5. This ensures that the computer is less swayed by moves that result in a loss, especially since the dynamics of Crazyhouse mean that just one wrong move will doom even hundreds of accurate ones. \\ 

\noindent Therefore, our output is an array of length 2308, for which the only non-zero entry can have a value of 0.1, 0.5, or 1.

\subsubsection{Value Representation}

For each position in a game, we attach a value of 1 for a win, 0 for a loss, or 0.5 of a draw depending on the final result of the position for the player to move.

\subsection{Training Dataset}

In recent years, Stockfish has reached a level of play higher than any human Crazyhouse player. Thus, instead of using human games, SixtyFour was trained on a total of 134,386 self-play games and roughly 11.2 million moves played by the multivariant version of Stockfish. In each position, Stockfish looks at a maximum of a million nodes and, unlike humans, plays until checkmate, in an attempt to ensure that mating patterns are learned by the computer. Furthermore, in order to ensure diversity, each game begins with one of the thousand most popular Crazyhouse openings played by humans, per the Lichess Opening Explorer.

\subsection{Model Architecture}

With the success of Residual Networks in later versions of AlphaGo\cite{SilverHuangEtAl16nature} and the generalized AlphaZero \cite{silver2017mastering1} \cite{silver2017mastering2}, we created a specialized ResNet architecture that outputs a policy (moves that are most likely to be played in a position)and value (an objective winning evaluation of a position) score for any board position. Since every small feature of the board is important to the outcome and decision making of the game, our model does not utilize any pooling layers, and the padding is specifically chosen to preserve the size of the image after each layer. Experiments with architectures of larger residual networks or pure convolutional neural networks did not attain a comparable mastery of the game; both overfitted the training dataset and took a significantly longer time to train.

As shown in Figure 2, the final residual network consists of 12 basic blocks. Each block consists of 256 kernels of size 3 $\times$ 3, batch normalization, and a rectified linear activation. The final fully connected layer is then connected to a policy head with 16 planes and a value head with 8 planes, each of which later pass through a softmax and tanh layer respectively to output the policy vector of dimension 2308 and a win probability from 0 to 1.  

\begin{figure}[!htb]
\centering
\includegraphics[width=11cm]{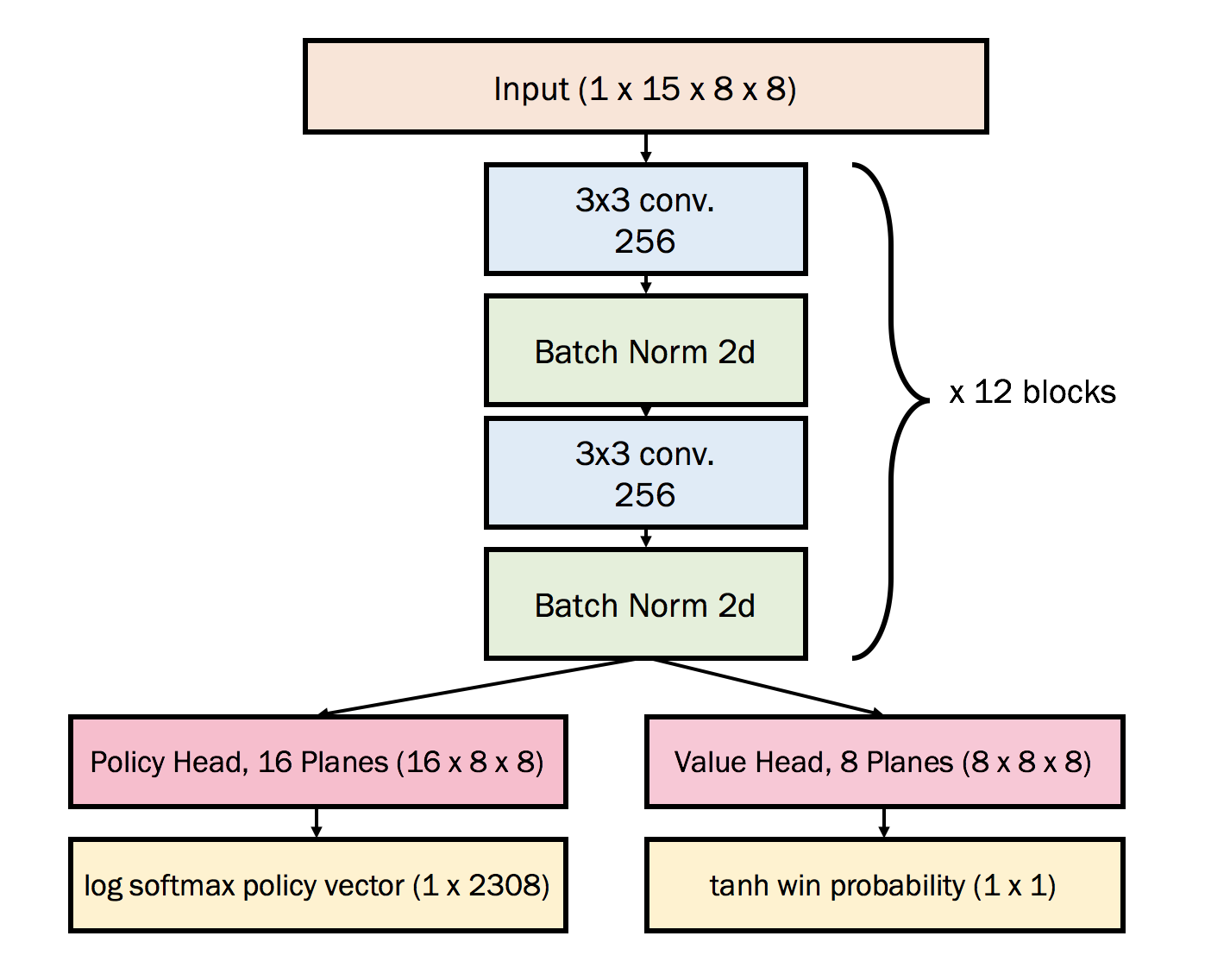}
\textbf{\caption{A diagram of the final residual network architecture}}
\label{network}
\end{figure}

\subsubsection{Training Hyperparameters}
For final results, we trained the above residual network on the full Stockfish self-play dataset with the use of an Adam optimizer and two loss functions. For the policy output, we implemented a Poisson NLL (negative log likelihood) loss function, chosen for its high accuracy in other classifier problems; for the value output, we chose a MSE (mean square error) loss function. This network was trained for a total of 8 epochs at an initial learning rate of 0.001. The results will be shown in the next section of the paper.

\subsection{Monte Carlo Tree Search}
We utilize a depth-first search algorithm, the Monte Carlo Tree Search, to traverse the state space. Instead of choosing random positions to explore, our neural network model guides this search by calculating the score, $V_{node}$, of each node in a position, and selecting the move with the highest score. The formula for this is shown in the equation below, and is a modification of the PUCT Algorithm.

\begin{equation} \label{eq:1}
\begin{split}
V_{node} = \frac{c(U_{node}+Noise(\alpha, \beta)) + W_{node}}{2}+ \gamma_{exploration} \frac{\sqrt{N_{parent}}}{1+n_{node}}
\end{split}
\end{equation}

In this equation, the number of visits of this position by the tree search is denoted as $N_{node}$, while the number of playout wins from this node is denoted as $W_{node}$. Checkmates are given a win score of 1.5 rather than the conventional 1.0 in order to incentivise the computer to choose lines that directly lead to mate. $c$ denotes a constant inversely proportional to the $\max(U_{nodes})$ score at any parent position. The noise function $Noise(\alpha, \beta)$ outputs a random number between $-\frac{\alpha}{x\beta}$ and $\frac{\alpha}{x\beta}$, where $x$ represents the number of moves already played in the game. We define $\alpha$ to be the noise constant and $\beta$ to be the decay value. Finally, $\gamma_{exploration}$ represents an arbitrary exploration constant, which we set to 0.5 for our agent.
\\ 

For each move chosen from the root position, the search algorithm will explore a subvariation of the move until it reaches a decisive result or a certain specified depth. If the game does not end, then the value evaluation of the neural network in the final position is used as the win probability of the variation. The result of this playout will be used to update the scores of each move. This can be repeated for as many times before computer decides to make an actual move.

\subsection{Time Management}

SixtyFour utilises a simple algorithm in which it uses roughly 10\% of its remaining time in the first fifteen plies and roughly 13.33\% ($\frac{2}{15}$) of its remaining time for every move thereafter. When under ten seconds, the program will disregard the tree search and output the first move
\\ \\
In addition to just time management, our program chooses a different depth for its tree search depending on how much time is allocated on the clock. If the program has more than fifteen minutes to make a move, each playout will search a variation of 20 plies. Otherwise, if the program has between five to fifteen minutes to make a move, each playout will search a variation of 15 plies. This number decreases further to 12 if there is between two to five minutes left on the clock, 10 if there is between one to two minutes left on the clock, and 4 if there is less than one minute of the clock.

\begin{table}[!htb]
\centering
\begin{tabular}{ |p{5cm}|p{4cm}|}
\hline
\multicolumn{2}{|c|}{\textbf{Search Depth vs Time on Clock}} \\
\hline
\textbf{Seconds left on the clock} & \textbf{Search Depth} \\
\hline
$\geq$750 & 20 \\
300-750 & 15 \\
120-300 & 12 \\
60-120 & 10 \\
10-60 & 4 \\
$\leq$10 & 0 \\

\hline
\end{tabular}
\textbf{\centering \caption{The Search Depth of SixtyFour depends on the remaining time on the clock. The tree search is disabled if only ten seconds or less are left on the clock.}}
\end{table}

\section{Results}

\subsection{Policy Prediction Accuracy}

To test the strength of our engine, we evaluate the performance of the residual network on a validation dataset of 500,000 moves played by the multivariant version of Stockfish. Our final run yielded a prediction accuracy of around 54.1\%, comparable to the 55.4\% accuracy attained by the first version of the Go-playing neural network program, AlphaGo, and much higher than the 44.4\% accuracy rate achieved by neural network models prior to AlphaGo. \cite{SilverHuangEtAl16nature} Moreover, compared to the initial policy loss of 2.564 and the initial value loss of 1.453, our model obtained a final policy loss of 0.821 and final value loss of 0.376 at the end of its training. The diagram on the next page shows the improvement of the neural network during a span of eight epochs.

\begin{center}
\begin{tikzpicture}
\begin{axis}[
    title={Validation Accuracy Over 8 Epochs},
    xlabel={Epochs},
    ylabel={Accuracy [\%]},
    xmin=0, xmax=8,
    ymin=0, ymax=60,
    xtick={0,1,2,3,4,5,6,7,8},
    ytick={0,10,20,30,40,50,60},
    legend pos=north west,
    ymajorgrids=true,
    grid style=dashed,
]
 
\addplot[
    color=blue,
    mark=square,
    ]
    coordinates {
    (0,0)(1,36.8)(2,42.1)(3,46.7)(4,48.2)(5,50.3)(6,52.2)(7,53.3)(8,54.1)
    };
    \legend{CrazyhouseResNet}
]
\end{axis}
\end{tikzpicture}
\end{center}

\subsection{Self-Play Games}

\subsubsection{Raw Policy Network without Tree Search}

We first show sample games of our Crazyhouse Chess engine played without the initialization of a tree search. We tune the noise constant as 0.3 and a decay value of 6. The average length of a game was 65.02 plies, slightly longer than the average online game. (For the months of May and June, 2018, the average game length for players with a $\geq$2000 rating on lichess.org was 59.13)  Many of the self-play games show tremendous promise for the engine and its intuition when launching an attack against the king. Below, we select four of the most accurate games that show the program and its aggressive play style.
\\ \\
A near perfect game is occasionally played when the computer plays a common opening and reaches a position similar or, in extreme cases, identical to the training data. This can be verified by the quality of the moves, as well as a statistical metric, the average centipawn loss (ACPL), for the following self-play examples. \\

\newpage

\textbf{Game 1:} \\
{1. Nf3 Nc6 2. d4 Nf6 3. Nc3 d5 4. Bf4 Bg4 5. e3 e6 6. Be2 Bb4 7. O-O Bxc3 8. bxc3 N@e4 9. Ne5 Nxc3 10. Nxc6 bxc6 11. N@e4 Nfxe4 12. f3 P@f2+ 13. Kh1 Qh4 14. B@g3 Qh3 15. Qe1 Qxg2+ 16. Kxg2 P@h3+ 17. Kh1 fxe1=Q 18. Raxe1 Q@g2\# 0-1
} \\ 
\textbf{White ACPL: 105 Black ACPL: 46} \\

\begin{figure}[!htb]
\centering
\includegraphics[width=5cm]{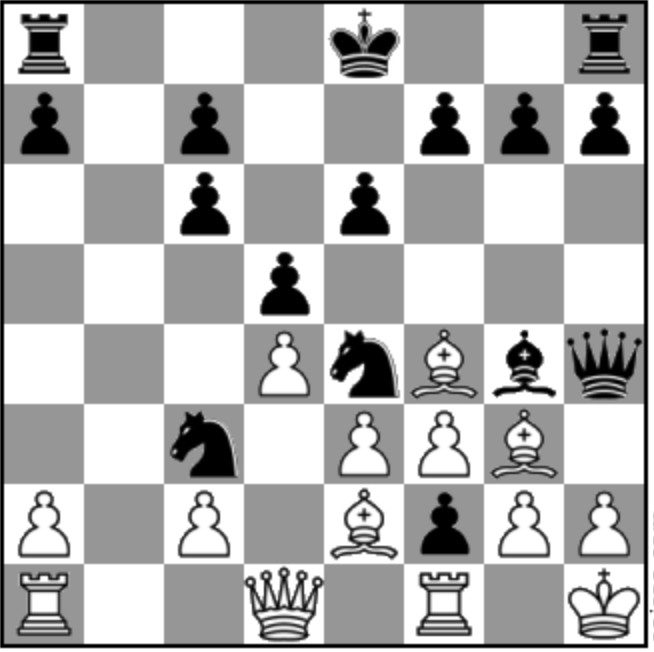}
\textbf{\caption{Black to move. White has no pieces in its pocket whereas Black has two knights. After B@g3, Black chooses to play Qh3!!. White plays Qe1, and now Black gets to checkmate the Black king after 15.... Qxg2!! 16. Kxg2 P@h3 17. Kh1 fxe1=Q 18. Raxe1 Q@g2\#}}
\label{qh3}
\end{figure}

\textbf{Game 2:} \\
{1. e4 e5 2. Nf3 Nc6 3. Bc4 Bc5 4. O-O Nf6 5. Nc3 O-O 6. d3 d6 7. Bg5 h6 8. Bh4 Bg4 9. Nd5 Nxd5 10. Bxd5 N@f6 11. c3 Nxd5 12. Bxd8 Raxd8 13. exd5 Bxf3 14. Qxf3 Nd4 15. cxd4 B@g4 16. Qxg4 N@f6 17. B@f5 Bxd4 18. B@e6 B@g6 19. Bxf7+ Rxf7 20. P@e6 Bxf5 21. Qxf5 B@g6 22. N@e7+ Kh8 23. Nxg6+ Kg8 24. N@e7+ Kh7 25. Nf8+ Kh8 26. N@g6\# 1-0
} \\
\textbf{White ACPL: 40 Black ACPL: 78}\\

\begin{figure}[!htb]
\centering
\includegraphics[width=5cm]{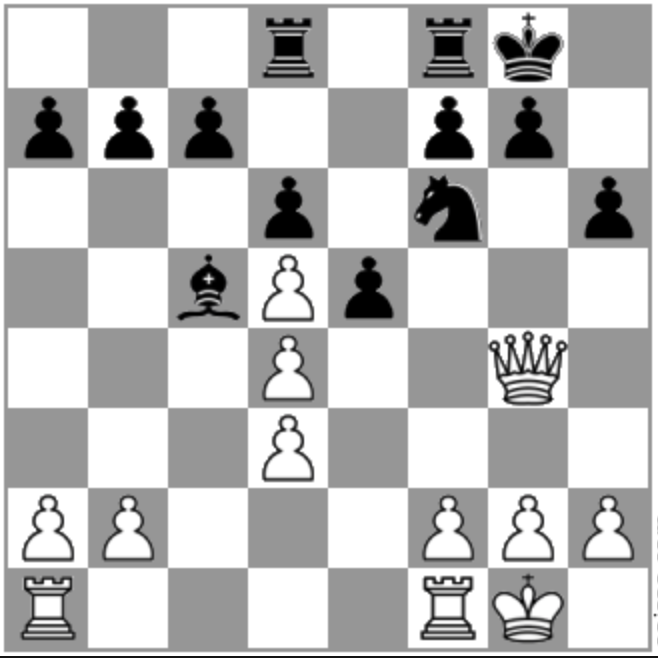}
\textbf{\caption{Game 2, White to move. White has three knights, a bishop, and a queen in its pocket. Black has a bishop in its pocket. Black has just played N@f6, threatening to capture the queen on the next move. Instead of moving out of danger, White plays B@f5!! And the queen cannot be captured or else Q@h7\# will follow. White converts this into a victory nine moves later, as seen in Figure 5.}}
\label{b@f5}
\end{figure}

\begin{figure}[!htb]
\centering
\includegraphics[width=5cm]{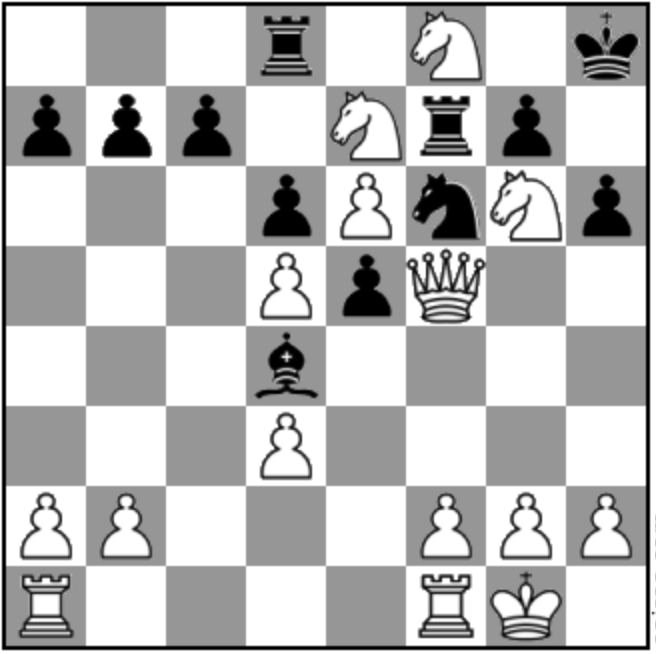}
\textbf{\caption{Final position of Game 2.}}
\label{game3mate}
\end{figure} 

\newpage 

\textbf{Game 3:} \\
{1. e4 e5 2. Nf3 Nc6 3. Bc4 Bc5 4. O-O Nf6 5. d3 O-O 6. Bg5 d6 7. Nbd2 Bg4 8. h3 Bh5 9. c3 h6 10. Bh4 Bxf3 11. Nxf3 N@f5 12. exf5 e4 13. dxe4 Ne5 14. Nxe5 dxe5 15. Bxf6 Qxf6 16. N@d5 Qd8 17. Qg4 B@e6 18. Qxg7+ Kxg7 19. N@h5+ Kh7 20. N@f6+ Kh8 21. P@g7\# 1-0
} \\
\textbf{White ACPL: 45 Black ACPL: 91} \\

\textbf{Game 4:} \\
{1. e4 e5 2. Bd3 Nc6 3. Nf3 d6 4. O-O Be7 5. Nc3 Nf6 6. Nxe5 Nxe5 7. P@e6 Bxe6 8. Qe2 Nxd3 9. cxd3 N@d4 10. N@h5 Nxe2+ 11. Nxe2 O-O 12. N@e5 dxe5 13. Neg3 P@h3 14. gxh3 Q@g2+ 15. Kxg2 Bxh3+ 16. Kxh3 B@g4+ 17. Kh4 P@g5+ 18. Kxg5 N@d4 19. P@h6 gxh6+ 20. Kh4 P@g5\# 0-1
} \\
\textbf{White ACPL: 87 Black ACPL: 36} \\

Each of these games feature a $\leq$ 100 ACPL score for both sides, which demonstrates the ability of the engine to maintain its advantage from the opening without any mistakes. 

\subsection{Games against Humans}

In addition to self-play games, a Lichess account, "SixtyFourEngine", was used to test the results of our network. In a total of 294 rated games, it has reached a top rating of 2278, a rating higher than 97\% of all Crazyhouse users on lichess.org. We showcase a few of its best games below. \\

\noindent \textbf{Game 1: 0.5m+0s, SixtyFourEngine vs mathace} \\
1. e4 Nc6 2. Nc3 e5 3. Nf3 Nf6 4. Bc4 Be7 5. Bxf7+ Kxf7 6. P@d5 Nd4 7. Nxe5+ Kg8 8. Nf3 Nxf3+ 9. Qxf3 N@d4 10. Qd3 B@b6 11. O-O Bec5 12. P@e3 Nxc2 13. Qxc2 d6 14. d3 P@f7 15. Bd2 Bg4 16. h3 Bh5 17. N@f5 Qd7 18. d4 Bb4 19. e5 Bxc3 20. exf6 Bxd2 21. N@e7+ Kf8 22. fxg7+ Ke8 23. gxh8=Q+ B@f8 24. N@f6+ Kd8 25. Nxd7 Kxd7 26. P@c6+ bxc6 27. dxc6+ Ke6 28. P@d5\# 1-0

\vspace{0.5cm}

\noindent \textbf{Game 2: 0.5m+0s, SixtyFourEngine vs mathace} \\
1. e4 Nc6 2. Nf3 Nf6 3. Nc3 e5 4. Bc4 Be7 5. O-O O-O 6. d3 d6 7. Ng5 Nd4 8. Be3 Qe8 9. f4 Ng4 10. fxe5 Bxg5 11. Bxd4 N@e3 12. Qf3 Nxc4 13. dxc4 dxe5 14. Bf2 Nxf2 15. Rxf2 B@d4 16. Nd5 Bxf2+ 17. Qxf2 B@d4 18. P@e3 Bgxe3 19. Nxe3 P@f4 20. B@d2 Bxe3 21. Bxe3 fxe3 22. Qxe3 B@d4 23. B@f2 Bxe3 24. Bxe3 P@f4 25. N@f6+ gxf6 26. N@h6+ Kh8 27. B@g7+ Kxg7 28. N@h5+ Kg6 29. B@f5+ Kxh5 30. P@g4+ Kg5 31. h4+ Kxh4 32. Bf2+ Kg5 33. Kh1 Kxh6 34. Bh4 Kg7 35. Bxf6+ Kh6 36. P@g5\# 1-0

\vspace{0.5cm}

\noindent \textbf{Game 3: 1m+2s, okei vs SixtyFourEngine} \\
1. d4 d5 2. Nc3 Bf5 3. Bf4 e6 4. e3 Bd6 5. Bb5+ Nc6 6. Nge2 Ne7 7. O-O O-O 8. Bxc6 bxc6 9. N@e5 f6 10. Ng3 fxe5 11. dxe5 Bxe5 12. Bxe5 P@f6 13. Nxf5 Nxf5 14. Bg3 Nxg3 15. fxg3 N@f5 16. Rxf5 exf5 17. B@e6+ B@f7 18. Bxf5 B@g6 19. B@h3 Bxf5 20. Bxf5 B@g6 21. B@h3 Bxf5 22. Bxf5 B@g6 23. Bxg6 Bxg6 24. B@h3 N@f5 25. P@f2 B@e6 26. N@f4 Qe7 27. Nxe6 Qxe6 28. N@d4 Nxd4 29. Bxe6+ Nxe6 30. B@b7 Rad8 31. Bxc6 B@e4 32. P@f3 Bxf3 33. gxf3 N@h3+ 34. Kg2 N@f4+ 35. exf4 Nhxf4+ 36. gxf4 Nxf4+ 37. Kg1 Nh3+ 38. Kf1 R@g1+ 39. Ke2 Nf4+ 40. Ke3 d4+ 41. Kxf4 P@g5\# 0-1

\vspace{0.5cm} 

\noindent \textbf{Game 4: 5m+7s, SixtyFourEngine vs IQ\_QI} \\
1. Nf3 d5 2. d4 Bf5 3. Bf4 Nf6 4. e3 e6 5. Be2 Bd6 6. Bxd6 cxd6 7. O-O O-O 8. Nc3 Nc6 9. B@h4 B@e7 10. a3 a6 11. Bd3 Bg6 12. Bxg6 hxg6 13. B@g5 B@h5 14. h3 Qd7 15. Qd3 Rac8 16. Nh2 e5 17. Bxf6 Bxf6 18. Bxf6 gxf6 19. Nxd5 Qe6 20. N@b6 B@e2 21. Qxe2 Bxe2 22. B@g4 Bxg4 23. Nxg4 B@h5 24. Ndxf6+ Kg7 25. B@h6+ Kh8 26. P@g7\# 1-0

\vspace{0.5cm}

\noindent \textbf{Game 5: 2m+1s, SixtyFourEngine vs mathace} \\
1. e4 Nc6 2. Nc3 e5 3. Nf3 Nf6 4. Bc4 Be7 5. O-O O-O 6. d3 d6 7. Ng5 Nd4 8. Be3 Qe8 9. f4 Ng4 10. fxe5 Bxg5 11. Bxg5 Nxe5 12. B@e3 Ndc6 13. Nd5 Bg4 14. P@h6 Bxd1 15. Raxd1 N@g4 16. hxg7 Kxg7 17. Bf6+ Kg8 18. P@g7 {Black resigns.} 1-0

\vspace{0.5cm}

\noindent \textbf{Game 6: 2m+0s, SixtyFourEngine vs anticreeps} \\
1. d4 Nf6 2. Nf3 d5 3. Bf4 Bf5 4. e3 e6 5. Be2 Be7 6. Nh4 Be4 7. O-O O-O 8. Nd2 Nc6 9. Rc1 Bg6 10. Nxg6 hxg6 11. c3 Ne4 12. Nxe4 dxe4 13. B@c2 g5 14. Bg3 f5 15. N@g6 Rf7 16. Bb3 N@d5 17. Bh5 Rf6 18. c4 Nb6 19. c5 Nd5 20. Ne5 Nxe5 21. Bxe5 Rh6 22. N@f7 Qf8 23. Nxh6+ gxh6 24. Bxd5 N@h4 25. Bxe6+ N@f7 26. Bexf7+ Qxf7 27. Bxf7+ Kxf7 28. R@g7+ Ke6 29. Q@f7+ Kd7 30. Qxe7+ Kc6 31. Qxc7+ Kb5 32. Qb3+ Ka6 33. Qcxb7+ Ka5 34. Q3b4\# 1-0

\vspace{0.5cm}

\noindent \textbf{Game 7: 2m+1s, mchelken vs SixtyFourEngine} \\
1. e4 Nf6 2. e5 d5 3. d4 Ne4 4. Bd3 Bf5 5. Nf3 e6 6. Bxe4 dxe4 7. Nfd2 Nc6 8. c4 Nxd4 9. N@b3 P@c2 10. Qh5 B@g6 11. Qh4 cxb1=Q 12. Rxb1 N@d3+ 13. Kf1 Nxb3 14. axb3 Qxh4 15. P@g3 Qxh2 16. Rxh2 P@e2+ 17. Kxe2 Nxc1+ 18. Rxc1 B@d3+ 19. Ke3 N@g4+ 20. Kd4 Nxe5 21. P@d7+ Nxd7 22. N@e5 Nxe5 23. Kxe5 P@d6+ 24. Kf4 Q@e5+ 25. Ke3 P@d4\# 0-1

\subsubsection{Games against Stockfish}

\textbf{Game 8: 1m+0s, Stockfish AI Level 4 (Lichess) vs SixtyFourEngine} \\
1. e3 e5 2. Bc4 Nc6 3. Bxf7+ Kxf7 4. Nf3 d5 5. Nxe5+ Nxe5 6. Qh5+ N@g6 7. P@f5 Nf6 8. fxg6+ hxg6 9. Qxe5 Bd6 10. P@e6+ Bxe6 11. Qxe6+ Kxe6 12. N@f4+ Bxf4 13. N@c5+ Ke7 14. Nxb7 Bxh2 15. Rxh2 Rxh2 16. B@a3+ P@d6 17. B@f1 R@h1 18. Nxd8 Rxf1+ 19. Kxf1 Rh1+ 20. Q@g1 Rxg1+ 21. Kxg1 N@e2+ 22. Kh2 Ng4+ 23. Kh3 Q@h2+ 24. Kxg4 Q@h5\# 0-1

\vspace{0.5cm}

\noindent \textbf{Game 9: 1m+10s, SixtyFourEngine vs Stockfish AI Level 5 (Lichess)} \\
1. e4 d5 2. exd5 Nf6 3. Nc3 Bf5 4. Nf3 c6 5. Bc4 e6 6. dxe6 Bxe6 7. Bxe6 fxe6 8. O-O Na6 9. P@f7+ Kd7 10. d4 Kc8 11. P@e5 P@g4 12. Ng5 Nc7 13. exf6 Qxf6 14. B@e5 B@d6 15. Bxf6 gxf6 16. Nxe6 Bxh2+ 17. Kxh2 B@d6+ 18. P@e5 Nxe6 19. Q@e8+ N@d8 20. Qxg4 P@d7 21. exd6 P@f5 22. Qxf5 Bxd6+ 23. P@e5 P@g3+ 24. fxg3 Rf8 25. Qfxe6 dxe6 26. P@d7+ Kb8 27. exd6 b6 28. P@c7+ Kb7 29. c8=Q+ Rxc8 30. dxc8=Q+ Kxc8 31. B@a6+ Q@b7 32. R@a8\# 1-0

\vspace{0.5cm}

\noindent \textbf{Game 10: 5m+8s, SixtyFourEngine vs Stockfish AI Level 5 (Lichess)} \\
1. e4 d5 2. exd5 Qxd5 3. Nc3 Qa5 4. d4 P@e6 5. Nf3 a6 6. Bc4 b5 7. Bd3 Nd7 8. O-O Bb7 9. Re1 Ngf6 10. Bf4 b4 11. Ne4 Nh5 12. Bd2 Nhf6 13. Nfg5 g6 14. Nxf7 Nxe4 15. Nxh8 N@h3+ 16. gxh3 Ndf6 17. Bxe4 Nxe4 18. Rxe4 B@g8 19. N@c5 Qxc5 20. dxc5 N@d7 21. Bxb4 Bxe4 22. Qxd7+ Kxd7 23. N@e5+ Kc8 24. Q@d7+ Kb7 25. N@a5+ Ka7 26. Qxc7+ Bb7 27. Qxb7\# 1-0

\vspace{0.5cm}

\noindent \textbf{Game 11: 2m+10s, SixtyFourEngine vs Stockfish AI Level 5 (Lichess)} \\
1. e4 Nf6 2. Nc3 Nc6 3. Nf3 e5 4. Bc4 b5 5. Bxb5 Nd4 6. Nxd4 exd4 7. O-O Bd6 8. N@f5 O-O 9. P@h6 g6 10. Nxd6 cxd6 11. B@g7 N@e6 12. Bxf8 dxc3 13. Bxd6 a6 14. P@e7 Qe8 15. dxc3 axb5 16. R@f8+ Nxf8 17. exf8=R+ Qxf8 18. Bxf8 Nh5 19. P@g7 N@h3+ 20. gxh3 R@g2+ 21. Kxg2 Nf4+ 22. Bxf4 P@f3+ 23. Kg1 f6 24. Q@h8+ Kf7 25. g8=Q+ Ke8 26. B8d6+ N@f8 27. Qxf8\# 1-0

\newpage
\newpage
\section{Conclusion}

This project presents a novel engine that, with the help of a tree search, has reached a level comparable to a master in the game of Crazyhouse. The program reached a rating of 2278 at its peak, a skill level that puts it in the top 97\% percentile of all Crazyhouse players on the site of lichess.org. While this is a great accomplishment, the SixtyFour engine still has many weaknesses that we hope to improve upon in the near future.

\subsection{Improvements}

We hope to improve both the quality and quantity of the training data. Our current database does not include any games played by two players of drastically different skill levels; by adding and learning from the moves of the stronger player in such games, the neural network model should begin punish inaccuracies more accurately. 
\\ \\
Moreover, the current SixtyFour engine is slow in its overlooks mate threats and plays suboptimal moves in positions it needs to defend accurately in. There are two ways to overcome this weakness. First, we hope to increase the speed of the tree search to $>$12 NPS; this will allow the engine to calculate more variations and give a more concrete justification for a move, rather than blindly following the intuition of the neural network to be right at all times. Second, we hope to implement a more advanced time management algorithm that will better allocate its time to calculate positions of higher complexity and those in question.
\\ \\
Finally, we hope to implement generative adversarial networks to create sample board positions and move choices in the future.

\subsection{Practical Takeaways}

The game of Crazyhouse exemplifies many of the challenges in many real-life machine learning projects. In the world of developing self-driving cars, for example, a model will often find itself having to make split second decisions in unseen scenarios. The results of our Crazyhouse agent, SixtyFour, and its wins against humans without a search algorithm show that these decisions can be made before a catastrophe; the careful crafting of training data is sufficient to compensate for the lack of a search algorithm. In conclusion, our experimentation of a unique game and its complex state space leads towards the construction of a general neural network that is not only capable of predicting an action in the present, but also an outlook of what to avoid in the future.

\section{Acknowledgments}

I would like to thank mathace for not only his generous offer of a GTX 1080Ti to help out with the training of SixtyFour neural network, but also for his guidance and testing of the chess engine throughout the process. This final project would not have been possible without him.

\newpage
\bibliographystyle{IEEEbib}

\end{document}